\documentclass[journal]{IEEEtran}
\usepackage{cite}
\usepackage{hyperref}
\usepackage{comment}

\ifCLASSINFOpdf
   \usepackage[pdftex]{graphicx}
\else
   \usepackage[dvips]{graphicx}
\fi
\usepackage{amsmath,amssymb}

\usepackage{makecell}
\usepackage{multirow}
\usepackage{multicol}
\usepackage{xspace}
\usepackage{booktabs}
\usepackage{caption}
\usepackage[dvipsnames]{xcolor}
\usepackage{cuted}

\usepackage{amsthm}

\usepackage{dsfont}
\usepackage{pifont}
\newcommand{\cmark}{\textrm{\ding{51}}}%
\newcommand{\xmark}{\textrm{\ding{55}}}%
\usepackage{wrapfig}

\makeatletter
\DeclareRobustCommand\onedot{\futurelet\@let@token\@onedot}
\def\@onedot{\ifx\@let@token.\else.\null\fi\xspace}

\def\eg{\emph{e.g}\onedot} 
\def\ie{\emph{i.e}\onedot}

\makeatother

\newcommand{\beforesection}{\vspace{-0mm}} 
\newcommand{\aftersection}{\vspace{-0mm}}
\newcommand{\beforesubsection}{\vspace{-1mm}}
\newcommand{\aftersubsection}{\vspace{-1mm}}

\newcommand{\revision}[1]{\textcolor{black}{#1}}

\usepackage{url}

\begin{document}
%
\title{Toward Real-Time Surgical Scene Segmentation via a Spike-Driven Video Transformer with Spike-Informed Pretraining}

%
%
%

\author{Shihao~Zou,
        Jingjing~Li,
        Wei~Ji,
        Jincai~Huang,
        Kai~Wang,
        Guo~Dan,
        Weixin~Si,
        Yi~Pan
\thanks{S. Zou is with Shenzhen Institutes of Advanced Technology, Chinese Academy of Sciences, Shenzhen, China. (E-mail: sh.zou@siat.ac.cn)}
\thanks{J. Li is with University of Alberta, Edmonton, Canada. (E-mail: jingjin1@ualberta.ca)}
\thanks{W. Ji is with School of Medicine, Yale University, New Haven, US. (E-mail: wei.ji@yale.edu)}
\thanks{J. Huang is with Southern University of Science and Technology, jointly with Shenzhen University of Advanced Technology, Shenzhen, China. (E-mail: 12533480@mail.sustech.edu.cn)}
\thanks{K. Wang is with Nanfang Hospital Southern Medical University, Guangzhou, China. (E-mail: kaiwangsmu@163.com)}
\thanks{G. Dan is with School of Biomedical Engineering, Shenzhen University, Shenzhen, China. (E-mail: danguo@szu.edu.cn)}
\thanks{W. Si and Y. Pan are with Faculty of Computer Science and Control Engineering, Shenzhen University of Advanced Technology, Shenzhen, China. (E-mail: 
panyi@suat-sz.edu.cn, siweixin@suat-sz.edu.cn)}

}

%
%

\markboth{Journal of \LaTeX\ Class Files,~Vol.~14, No.~8, August~2015}%
{Shell \MakeLowercase{\textit{et al.}}: Bare Demo of IEEEtran.cls for IEEE Journals}
%



\maketitle


\begin{abstract}
\revision{Modern surgical systems increasingly rely on intelligent scene understanding to improve intra-operative safety and situational awareness, with surgical scene segmentation playing a fundamental role in fine-grained surgical perception. Although recent Artificial Neural Network (ANN) models, especially large foundation models, have achieved impressive accuracy, their high computational and energy demands often hinder deployment in resource-constrained operative environments. 
To address this challenge, we explore Spiking Neural Network (SNN) as a highly efficient paradigm for surgical intelligence. However, its performance in surgical scene segmentation remains constrained by sparse spike representations and limited annotated surgical data. We therefore propose \textit{SpikeSurgSeg}, the first spike-driven video Transformer for surgical scene segmentation. It preserves the real-time and energy-efficient advantages of SNN, while achieving competitive performance against most ANN models in data-scarce surgical scenarios. 
Specifically, we introudce a spike-informed pretraining strategy based on Masked AutoEncoding (MAE), where mask generation is guided by spike firing activity to better align with sparse spike representations, together with a layer-wise tube masking scheme that reduces information leakage and encourages contextual reasoning. To further strengthen semantic representation, we introduce multi-spectral knowledge distillation, which aligns teacher ANN and student SNN features in the frequency domain, where the mismatch between continuous activation patterns and spike-driven temporal representations can be effectively mitigated. Built on the pretrained SNN encoder, we further design a lightweight spike-driven segmentation head that produces temporally consistent predictions while preserving the low-latency advantage of spike-driven inference. 
Extensive experiments on EndoVis18 and our in-house SurgBleed dataset show that \textit{SpikeSurgSeg} achieves mIoU comparable to State-Of-The-Art (SOTA) ANN models while reducing inference latency by at least $8\times$. Notably, it delivers over $20\times$ speedup relative to most foundation-model baselines, highlighting its promise for real-time surgical scene segmentation in resource-constrained settings.
}
\end{abstract}


%
\IEEEpeerreviewmaketitle

\beforesection
\section{Introduction}\label{sec:introduction}
\aftersection






\begin{figure}[t]
    \centering
    \includegraphics[width=\columnwidth]{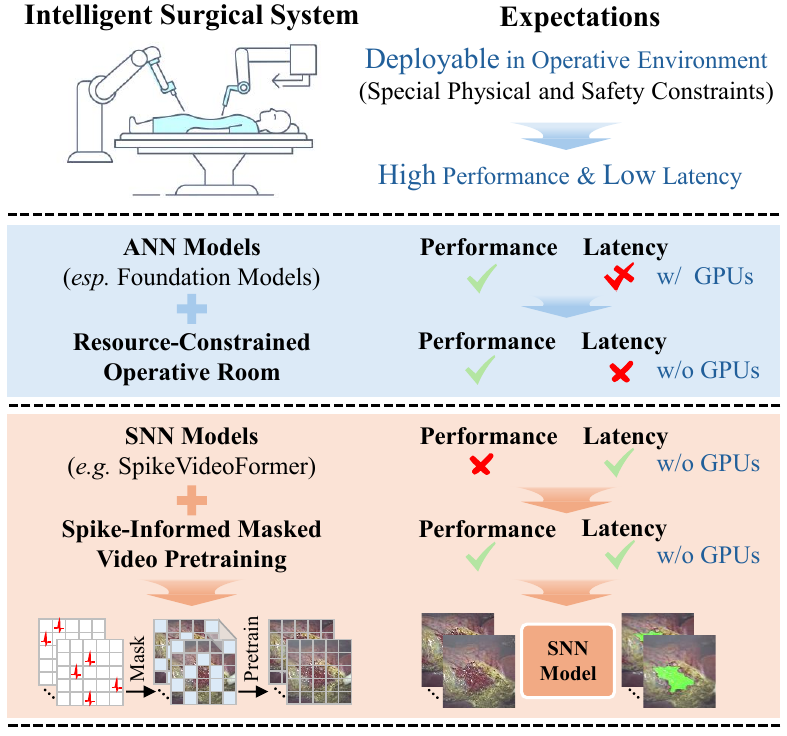}
    \vspace{-5mm}
    \caption{\textbf{Motivation of Our Work.} Intelligent surgical systems must achieve both high accuracy and low latency to be deployable in the operative environments. \revision{Although \textit{ANN models}, especially foundation models, deliver strong performance in surgical scene segmentation, they typically rely on power-hungry GPUs to maintain low latency, which is impractical under the physical and safety constraints of operative environments. In contrast, emerging \textit{SNN models} offer inherent efficiency but often suffer from performance degradation due to sparse spike representations. To address this limitation, we propose a spike-informed pretraining strategy that enhances surgical scene segmentation performance of SNN while preserving their low-latency advantage.}
    }
    \vspace{-4mm}
    \label{fig:concept}
\end{figure}

Modern surgery has transformed from an experience-driven art into a data-centric, technology-augmented discipline, propelled by the convergence of advanced imaging systems, robotic platforms, and multimodal intra-operative sensing technologies~\cite{qin2025medical,viviers2024advancing,long2025surgical,Kiyasseh2023SAIS}. These innovations have produced massive streams of heterogeneous surgical data, such as high-resolution surgical or endoscopic videos, laying the foundation for computational models to perceive and interpret surgical scenes. Within this paradigm, surgical scene segmentation has emerged as a fundamental component, aiming to comprehensively recognize and contextualize instruments, anatomical structures, and operative states~\cite{jin2022exploring,yue2024surgicalsam,liu2024surgical,maier2022surgical}.


\revision{Despite notable progress, surgical scene segmentation still faces major barriers to clinical deployment. Deep learning models, especially large foundation models with hundreds of millions of parameters, require substantial computational and energy resources, making real-time execution in the operating room impractical~\cite{maier2022surgical}. These challenges are further exacerbated by the physical and safety constraints of operative environments, \ie, limited space, strict heat dissipation requirements, and tight power budgets, which preclude the use of high-end GPU clusters (Fig.~\ref{fig:concept}). As a result, a persistent gap remains between research prototypes and deployable intelligent surgical systems: existing deep learning models struggle to simultaneously achieve high accuracy and low latency under energy-constrained hardware, hindering their seamless adoption in real-world surgical intelligent systems~\cite{carstens2023dresden}.}

To bridge this gap, we explore brain-inspired computing as a promising paradigm for building efficient and intelligent surgical systems. Within this paradigm, SNN stands out for its high energy efficiency by mimicking the temporal coding and asynchronous sparse spike communication of biological neurons~\cite{2025Neuromorphic,li2024brain}. Unlike conventional ANN, which relies on dense floating-point operations, SNN encodes information with discrete spikes and performs spike-driven, addition-only computations, thereby substantially reducing power consumption. Moreover, its intrinsic temporal dynamics naturally align with the sequential nature of surgical videos, making SNN a biologically inspired and computationally efficient solution for real-time surgical scene segmentation.

\revision{
However, directly applying SNN to surgical scene segmentation faces two key challenges. 1) Although SNN has achieved competitive performance with ANN on some generic vision tasks~\cite{fang2021deep,yao2023attention,yao2024spikev2,luo2024integer,lei2025spike2former,zou2025spikevideoformer}, existing architectures have been developed and evaluated primarily for static image benchmarks rather than dynamic video tasks. Consequently, their ability to capture the complex, domain-specific dynamics required for surgical video segmentation remains largely underexplored. 2) Annotated surgical datasets are typically limited in scale, making it difficult for SNN models--given their sparse, spike-driven nature--to achieve performance comparable to ANN counterparts. This makes it particularly challenging to preserve the efficiency of SNN while also ensuring strong performance under data-scarce surgical conditions.
}

\revision{
To address these challenges, we propose \textit{SpikeSurgSeg}, the first spike-driven video Transformer framework to tackle surgical scene segmentation. The proposed framework preserves the real-time capability and energy efficiency of SNN in source-constrained operative environments, while achieving highly competitive performance with most ANN models in data-scarce surgical scenarios. At its core is a \textit{spike-informed pretraining} strategy in which a spike-driven video Transformer is pretrained on unlabeled surgical videos to learn robust spatiotemporal representations prior to downstream segmentation finetuning. Unlike generic MAE-based masking schemes, our method uses spike firing activity to guide mask generation, enabling self-supervised SNN pretraining that is better aligned with sparse spike representations. We further introduce a layer-wise tube masking strategy within the SNN encoder to prevent information leakage and encourage contextual reasoning. To further strengthen semantic representation learning, we leverage a teacher ANN model to guide SNN pretraining. However, direct alignment between ANN and SNN features is inherently ill-posed due to the mismatch between continuous activation statistics and spike-based temporal accumulation. We therefore propose a multi-spectral knowledge distillation method that aligns features in the frequency domain, where surgical motion patterns and structural semantics are more stable and less sensitive to spatiotemporal perturbations.
}
\revision{
After pretraining, the SNN encoder is coupled with a lightweight spike-driven segmentation head that integrates a spike pyramid and a memory readout module. This design enables temporally consistent video segmentation while preserving the low-latency and low-energy advantages of spike-driven inference. Experiments show that our model achieves mIoU comparable to most ANN baselines, while delivering at least $8\times$ faster inference and $5.1\times$ lower energy consumption.
}

Our main contributions are summarized as follows:
\begin{itemize}
    \item \revision{We propose \textit{SpikeSurgSeg}, the first SNN framework to tackle surgical scene segmentation. The proposed framework preserves the real-time and energy-efficient advantages of SNN for resource-constrained operative environments and, with tailored spike-informed pretraining, achieves highly competitive performance against most ANN models in data-scarce surgical scenarios.}

    \item \revision{We introduce \textit{spike-informed pretraining} for the spike-driven video Transformer encoder. Unlike generic MAE-based masking schemes, we exploit spike firing activity to generate masks that better align with sparse spike representations. We further incorporate multi-spectral knowledge distillation to align teacher ANN and student SNN features in the frequency domain. Together, these designs produce more robust spatiotemporal representations for our SNN models and significantly improve downstream surgical scene segmentation performance.}

    \item We demonstrate that our SNN approach achieves mIoU comparable to SOTA ANN baselines on both the EndoVis18 dataset and our in-house SurgBleed dataset, while delivering \textit{at least} $8\times$ faster inference and $5.1\times$ lower energy consumption. Notably, it delivers \textit{over $20\times$ speedup} relative to most foundation-model baselines.
\end{itemize}

\beforesection
\section{Related Work}
\aftersection
\label{sec:related-work}

\textbf{Spiking Neural Network.} SNN is a brain-inspired learning framework that emulate the spike-based communication mechanism of the mammalian visual cortex~\cite{2025Neuromorphic,li2024brain}. Large-scale training of SNN remains challenging due to the non-differentiable nature of spike generation. Early approaches addressed this by converting trained ANN into SNN~\cite{deng2021optimal}, which achieved competitive performance on simple classification tasks but struggled with more complex vision problems. To overcome this limitation, direct training methods based on back-propagation through time (BPTT) with surrogate gradients~\cite{li2021differentiable,zhou2024direct} have been developed, enabling the design of increasingly powerful architectures. Among them, convolution-based SNN adopt deep CNN design principles to enhance representation capacity, where models such as SEW-SNN~\cite{fang2021deep} and MS-SNN~\cite{hu2024advancing} introduce spike- or membrane-level residual connections to alleviate vanishing gradients, although they remain limited on large-scale vision tasks. More recently, transformer-based SNN has advanced the field. Early spiking Transformers~\cite{yao2023attention} incorporated multi-dimensional attention but relied on real-valued membrane potentials, reducing efficiency. Subsequent works~\cite{zhou2022spikformer,yao2024spike} proposed spike-driven self-attention based on dot-product similarity between binary spike tensors, eliminating softmax and achieving linear complexity with respect to token length. Later developments include hybrid convolution–Transformer architectures such as SpikeFormer~\cite{yao2024spikev2}, masked self-supervised pretraining for spike-driven models~\cite{yao2025scaling}, and Hamming attention~\cite{zou2025spikevideoformer}, a theoretically grounded alternative to standard dot-product attention~\cite{vaswani2017attention}, further strengthening the foundation of spike-driven Transformer architectures.

\textbf{Masked Visual Modeling.} MVM has emerged as a dominant paradigm for self-supervised learning in vision, particularly for Transformer-based architectures. The seminal MAE framework~\cite{he2022masked} learns representations by reconstructing heavily masked image patches through masked token prediction. Its video extensions, such as VideoMAE and MAE-ST~\cite{tong2022videomae}, demonstrate that random spatiotemporal masking enables efficient and effective video representation learning, achieving large training speedups and strong performance. Beyond pixel reconstruction, methods like MaskFeat~\cite{wei2022masked} predict richer features such as HOG descriptors, improving generalization for video and image understanding. Subsequent works further extend the paradigm: MCMAE~\cite{gao2022mcmae} integrates masked convolutions with hybrid convolution–Transformer architectures and multi-scale supervision, while SparK~\cite{tian2023designing} adapts MVM to pure convolutional backbones using sparse encoders and hierarchical decoders. Other approaches explore improved training strategies, such as the student–teacher masked autoencoder SdAE~\cite{chen2022sdae} with multi-view masking for stronger downstream performance. At larger scale, DINOv2~\cite{oquab2023dinov2} demonstrates that discriminative self-supervised learning with ViTs can produce robust visual representations without reconstruction objectives. More recently, domain-specific designs such as CSMAE~\cite{shah2025csmae} introduce spatiotemporal importance-aware masking for surgical videos, achieving superior surgical step recognition compared with generic pretraining strategies.

\textbf{Surgical Scene Segmentation.} This task serves as a fundamental task in surgical scene understanding. Early approaches~\cite{zhao2020learning,jin2022exploring}, primarily based on CNNs and Transformers, focused on improving performance by incorporating spatial attention, optical flow, and motion priors. These methods gradually evolved from static-frame segmentation toward more sophisticated techniques that leverage intra- and inter-video relationships and contrastive learning. Building on the success of foundation models in natural image or video segmentation, \eg, SAM~\cite{kirillov2023segment} and SAM2~\cite{ravi2024sam2}, recent research has explored their adaptation to medical image segmentation. MedSAM~\cite{ma2024segment} develops a universal medical segmentation model based on SAM, trained on a large-scale medical image dataset. Other efforts~\cite{zhou2023text} finetune vision-language models for text-promptable surgical scene segmentation. More recent works~\cite{yue2024surgicalsam,paranjape2024adaptivesam,wu2025medical} propose efficient tuning strategies to incorporate surgical-specific knowledge, such as 3D medical images or surgical videos, into SAM for improving generalization. SurgicalSAM2~\cite{liu2024surgical} extends by finetuning SAM2 with a frame pruning mechanism, further enhancing inference speed.



\revision{
As noted in~\cite{maier2022surgical}, real-time performance is crucial for the clinical deployment of intelligent surgical systems. However, most existing approaches rely on ANN, particularly large foundation models, whose substantial computational and energy demands limit their use in resource-constrained operative environments. To address this challenge, we explore brain-inspired SNN as a highly efficient paradigm for surgical scene segmentation. Building on a spike-driven video Transformer with spike-informed pretraining, our framework learns robust surgical representations while preserving the low-latency and energy-efficient advantages of spike-driven inference.
}


\beforesection
\section{Preliminary}
\aftersection
\label{sec:snns}

\revision{
\textbf{Spiking Neuron Model.} The neuron model is the fundamental difference between the ANN and SNN. The \textit{Leaky Integrate-and-Fire neuron} (LIF) is a widely used computational unit in SNN. The LIF neuron maintains a membrane potential $u_{t}$, decay through time with a leaky constant $\beta$. The potential will be updated upon receiving input spike trains $X_{t}$ from its connected neurons over $T$ time steps. When the potential surpasses the threshold $V_{\text{th}}$ at time $t$, the neuron will emit a spike $S_{t}$ and undergoes a soft reset~\cite{zhou2022spikformer} by reducing its potential by $u_{\text{th}}$. The dynamics can be expressed as
\begin{gather}
    \label{eq:neuron}
    H_{t} = \underbrace{\beta U_{t-1}}_{\text{leak}} + \underbrace{X_{t}}_{\text{charge}}, \\
    S_{t} = \underbrace{\Theta (H_{t} - u_{\text{th}})}_{\text{spike}}, \quad U_{t} = H_{t} \underbrace{- u_{\text{th}}S_{t}}_{\text{reset}}, 
\end{gather}}
where $\Theta$ is the element-wise Heaviside step function:
\begin{equation}
    \Theta(H_{t} - u_{\text{th}}) =
    \begin{cases}
    1, & \text{if } H_{t} \geq u_{\text{th}}\\
    0, & \text{otherwise}
    \end{cases}. \nonumber
\end{equation}
For convenience, we denote the temporal spiking process of LIF neuron as $\mathcal{SN}_s(U)$ with scaled threshold $s\cdot u_{\text{th}}$.

\textbf{Feedforward Process.} SNN often consists of stacked layers of spiking neurons. For the $l$-th layer with $N^{(l)}$ neurons, we denote the membrane potentials and spike outputs at time $t$ as vectors $\boldsymbol{U}^{(l)}_{t} \in \mathbb{R}^{N^{(l)}}$ and $\boldsymbol{S}^{(l)}_{t} \in \{0, 1\}^{N^{(l)}}$, respectively. The synaptic weights between layers $l-1$ and $l$ are represented by $\boldsymbol{W}^{(l)} \in \mathbb{R}^{N^{(l)} \times N^{(l-1)}}$. Thus, the feedforward is defined as
\begin{gather}
    \label{eq:feedforward}
    \boldsymbol{H}_{t}^{(l)} = \beta \boldsymbol{U}^{(l)}_{t-1} + \boldsymbol{W}^{(l)} \boldsymbol{S}^{(l-1)}_{t}, \\
    \boldsymbol{S}^{(l)}_{t} = \Theta (\boldsymbol{H}_{t}^{(l)} - u_{\text{th}}), \quad
    \boldsymbol{U}^{(l)}_{t} = \boldsymbol{H}_{t}^{(l)} - u_{\text{th}}\boldsymbol{S}^{(l)}_{t}.
\end{gather}

\begin{figure*}[t]
    \centering
    \includegraphics[width=0.97\textwidth]{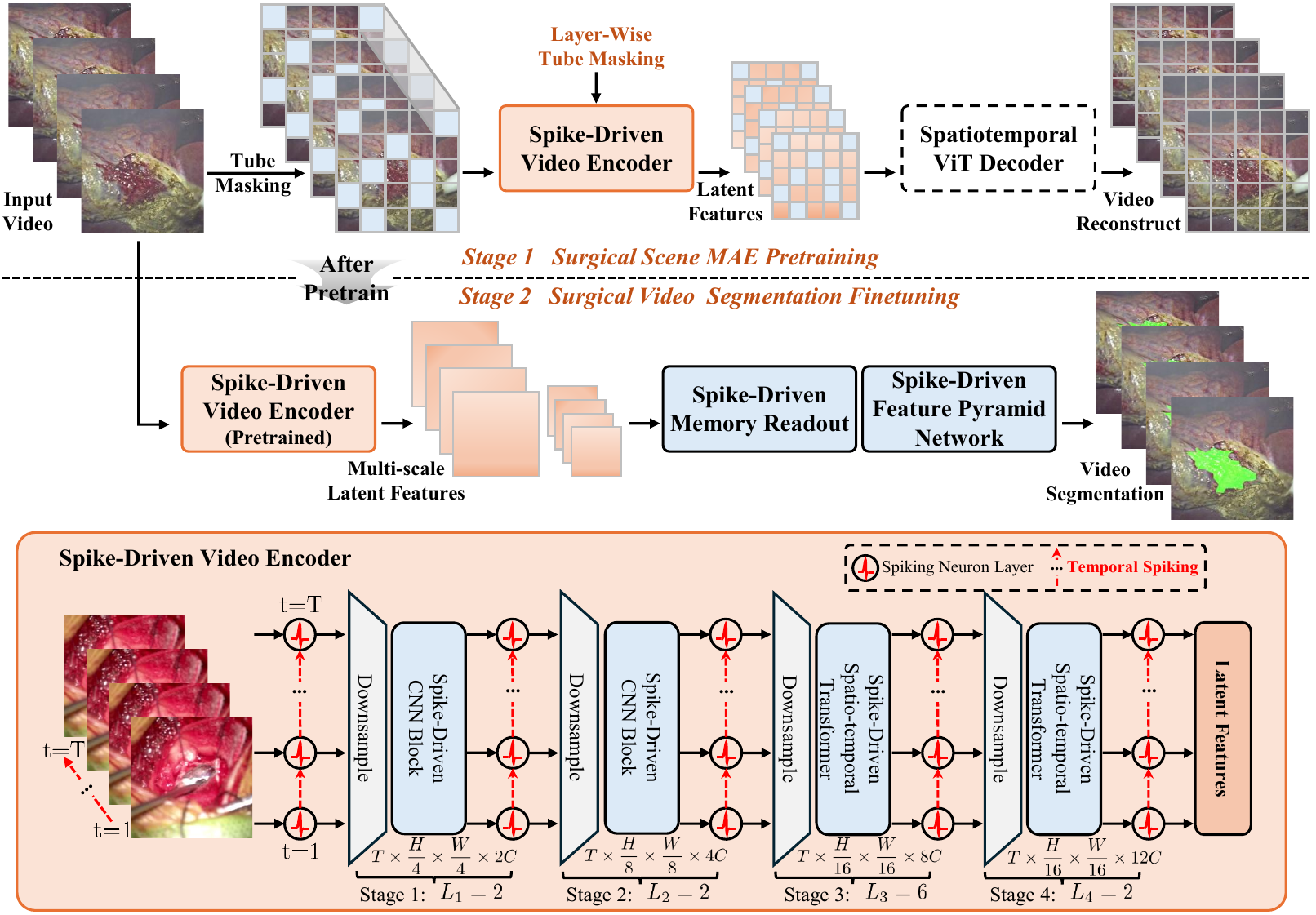}
    \vspace{-1mm}
    \caption{\textbf{SpikeSurgSeg Pipeline.} \revision{1) \textit{Stage 1}: The spike-driven video Transformer encoder is pretrained using a spike-informed strategy tailored for SNN based on MAE, where a spike-guided mask is generated and applied to the input video clip, and the masked regions are reconstructed from the unmasked context. 2) \textit{Stage 2}: Spike-driven memory readout and feature pyramid network are integrated with the pretrained spike video encoder, yielding a fully spike-driven SNN model that is then finetuned for downstream surgical scene segmentation. 3) The bottom block illustrates the encoder architecture, which consists of two spike-driven CNN blocks and two spike-driven spatiotemporal Transformer blocks featuring linear space–time complexity.}
    }
    \vspace{-3mm}
    \label{fig:pipeline}
\end{figure*}



\textbf{Efficiency of SNN.} The efficiency is largely attributed to its computational complexity and energy consumption.
\begin{enumerate}
    \item \revision{\textbf{\textit{SNN involves only simple additions during feedforward propagation}}, unlike ANN, which relies on costly matrix multiplications. Because the spike vector $\boldsymbol{S}^{(l-1)}_{t}$ is binary, the computation of $\boldsymbol{W}^{(l)} \boldsymbol{S}^{(l-1)}_{t}$ in Eq.~(\ref{eq:feedforward}) can be realized using additions only, leading to substantially higher efficiency, especially on neuromorphic hardware~\cite{2025Neuromorphic,2024Spike}.} For 45nm CMOS technology~\cite{horowitz20141}, a 32-bit floating-point multiply-and-accumulate (MAC) operation in ANN consumes $E_{\text{MAC}} = 4.6pJ$, while a simple accumulate (AC) operation in SNN requires only $E_{\text{AC}} = 0.9pJ$, a 5.1$\times$ energy reduction. Similarly, on the AMD Xilinx ZCU104 platform~\cite{li2023firefly}, SNN achieves 5529.6 AC GFLOPs per second, compared to 691.2 MAC GFLOPs for ANN, an 8$\times$ speedup.
    \item \revision{\textbf{\textit{SNN also benefits from sparse activation.}} Because most entries in $\boldsymbol{S}^{(l-1)}_t$ are zero, many AC operations in $\boldsymbol{W}^{(l)} \boldsymbol{S}^{(l-1)}_t$ can be skipped, thereby substantially reducing computational cost.} Assuming a spiking rate of $\rho$ (typically $\sim$20\%) over $T$ time steps in the $l$-th layer, the computational complexity decreases from $\mathcal{O}(T N^{(l-1)} N^{(l)})$ for ANN to $\mathcal{O}(\rho T N^{(l-1)} N^{(l)})$, yielding a $\rho\times$ reduction in computations.
\end{enumerate}



\revision{
\textbf{Integer Leaky Integrate-and-Fire Neuron.} IntLIF model was introduced in recent studies~\cite{luo2024integer,lei2025spike2former} which enhances the performance of SNN on complex vision tasks while incurring minimal additional computational cost. Unlike LIF neurons that produce binary 0/1 outputs, IntLIF neurons output integer-valued spikes. Specifically, an IntLIF neuron with integer setting $Z$ generates spike outputs within the range $S_{t} \in \{0, 1, \dots, Z-1\}$. While the training of IntLIF-based models follows the standard procedure, the inference process can be unfolded into $Z$ sub-steps of 0/1 spikes. In each sub-step, Eq.~(\ref{eq:feedforward}) still involves only additions, avoiding any costly multiplications. This IntLIF configuration is also adopted in the segmentation head of our approach to further improve model efficiency and performance.
}

\beforesection
\section{Method}
\aftersection

\beforesubsection
\subsection{Spike-Driven Video Encoder}
\aftersubsection
\label{sec:spikevideoformer}

\revision{
We adopt a spike-driven encoder based on a \textit{Conv+ViT} design, known for strong scalability and generalization in both ANN~\cite{yu2022metaformer} and SNN~\cite{zou2025spikevideoformer,yao2024spikev2}. As illustrated in Fig.~\ref{fig:pipeline}, the encoder combines spike-driven CNN blocks for efficient local feature extraction with spike-driven spatiotemporal Transformer blocks that model global dependencies via multiplication-free attention with linear space–time complexity, making it well suited for real-time surgical scene segmentation in resource-constrained settings. As spike-driven video Transformer~\cite{zou2025spikevideoformer} remains the only available SNN encoder designed for video modeling, we adopt it as our backbone without further comparison to alternative SNN architectures.
}

Specifically, the encoder takes a video clip $\boldsymbol{I}\in \mathbb{R}^{T \times H \times W \times 3}$ as input. The clip is processed by two spike-driven CNN blocks followed by two spike-driven spatiotemporal Transformer blocks, with downsampling modules inserted between stages to progressively reduce spatial resolution. At the $l$-th intermediate block ($l\in\{0,1,...,4\}$), the spatial dimension is $\frac{H}{2^l}\times \frac{W}{2^l}$. The final feature representation is denoted as $\boldsymbol{U}^{\text{latent}} \in \mathbb{R}^{T \times \frac{H}{16} \times \frac{W}{16} \times 12C}$, where $C$ is a tunable hyperparameter controlling the model capacity.



\revision{\textbf{Encoder Variations.} We use a compact model with $C=32$ and a larger model with $C=64$, containing 16.0M and 56.3M parameters, respectively. Both encoders are initially trained on the ImageNet1K dataset to learn general visual features, and the resulting weights are then used to initialize subsequent pretraining on surgical scenes. Architecture details are provided in Supplementary A.}

\beforesubsection
\subsection{Spike-Informed Pretraining for Surgical Scenes}
\aftersubsection
\label{sec:spike-transformer}


\revision{
Self-supervised pretraining has achieved remarkable success in ANN-based visual foundation models~\cite{he2022masked,gao2022mcmae,shah2025csmae,tong2022videomae,wang2023videomae,tian2023designing}, often yielding representations superior to those learned by direct task supervision. This is particularly valuable for surgical scene understanding, where annotated data are limited and expensive. It is even more critical for SNN, whose sparse spike representations often require more training data to achieve strong performance, motivating our spike-informed pretraining for data-scarce surgical video learning.
}

\revision{
\textbf{Spike-Informed Mask Generation.} SNN representations are inherently sparse, consisting of binary or integer spikes with a large proportion of inactive states (\ie, 0). Unlike dense ANN features, this spike-driven property provides a natural and interpretable measure of information content: regions with higher firing rates typically carry more informative content, whereas low-activity regions are often more redundant. Motivated by this observation, we propose a spike-informed tube masking strategy for SNN video pretraining, which leverages spike activity to guide masking. In contrast, existing work~\cite{yao2025scaling} explores conventional masked image pretraining for SNN without explicitly leveraging its spike-driven characteristics.}



\begin{figure}[!ht]
    \centering
    \includegraphics[width=\columnwidth]{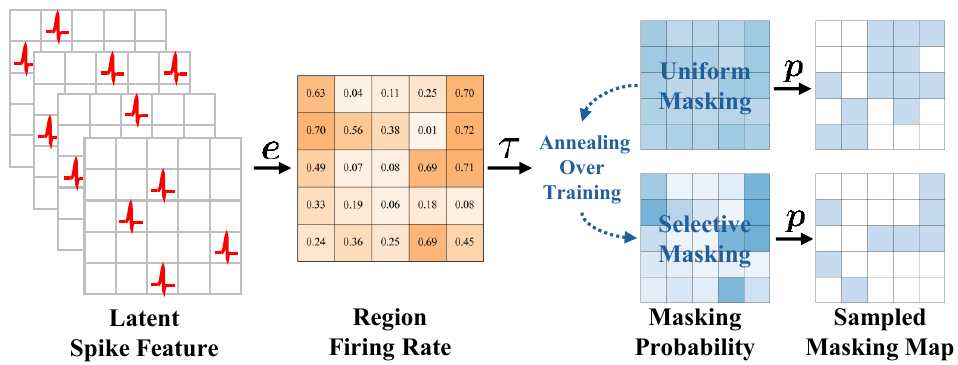}
    \vspace{-4mm}
    \caption{\revision{\textbf{Spike-Informed Mask Generation.} Masking is guided by spike activity, with higher firing rates indicating more informative regions. Annealing progressively shifts the sampling from uniform to selective to produce the final masking map.}   
    }
    \label{fig:spike-mask-gen}
\end{figure}

\revision{
Given the unmasked video, the encoder first produces latent feature maps $\boldsymbol{U}^{\text{latent}}$ and corresponding spiking features $\boldsymbol{S}^{\text{latent}}$. For each spatial location $(i,j)$, we compute the firing rate $\boldsymbol{e}_{ij}$ by averaging spike responses over the temporal and channel dimensions as the local information score:
\begin{gather}
    \boldsymbol{e}_{ij}=\frac{1}{T\cdot 12C}\sum_t\sum_c \boldsymbol{S}^{\text{latent}}_{tijc},
\end{gather}
where larger firing rate indicates richer spike dynamics and higher information content (see Fig.~\ref{fig:spike-mask-gen}). To decide masking regions, we define the masking probability as
\begin{equation}
    \boldsymbol{p}_{ij} = \frac{\exp(-\boldsymbol{e}_{ij}/\tau)} {\sum_{uv}\exp(-\boldsymbol{e}_{uv}/\tau)},
\end{equation}
where $\tau$ controls the sharpness of the distribution. It is initialized to a high value and gradually annealed over epochs, allowing mask generation to transition from uniform to more selective. Mask regions are randomly sampled according to $\boldsymbol{p}$ until a cumulative probability of 0.5 is reached, yielding the spike-informed 2D masking map $\boldsymbol{M}^{\text{latent}} \in \mathbb{R}^{\frac{H}{16} \times \frac{W}{16}}$.
}

\begin{figure}[!ht]
    \centering
    \includegraphics[width=\columnwidth]{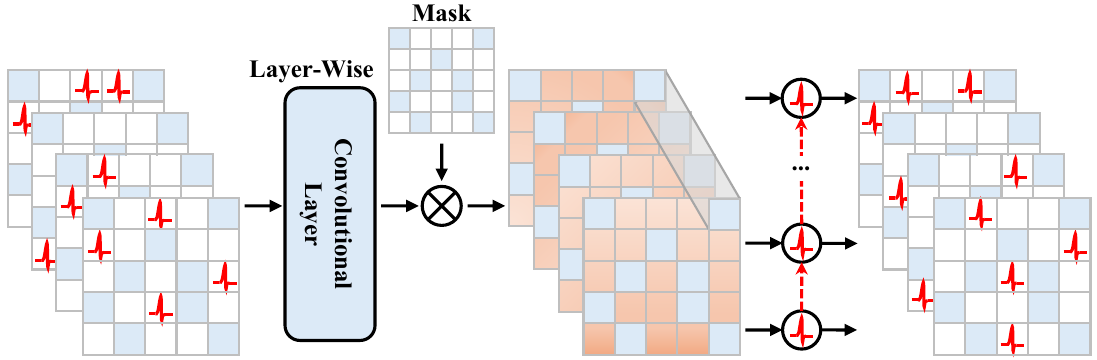}
    \vspace{-4mm}
    \caption{\textbf{Layer-Wise Tube Masking.} The tube-shaped mask is applied at each encoder layer to prevent any spike activity within the masked areas, ensuring that no spatial or temporal information leaks into the reconstruction process.
    }
    \label{fig:tube-masking}
\end{figure}

\textbf{Layer-Wise Tube Masking.} We further adopt layer-wise tube masking to mitigate distribution shifts and prevent information leakage that may occur when convolution or attention operations are applied directly to partially masked inputs~\cite{tian2023designing,yao2025scaling}. To obtain the tube masking map for the $l$-th intermediate block, $\boldsymbol{M}^{\text{latent}}$ is first repeated along the temporal dimension $T$ and then spatially expanded to cover patches of $2^{4-l} \times 2^{4-l}$ pixels. The resulting layer-wise tube masking map, $\boldsymbol{M}^l \in \{0,1\}^{T \times \frac{H}{2^l} \times \frac{W}{2^l}}$, matches the spatial resolution of the corresponding intermediate feature map. As illustrated in Fig.~\ref{fig:tube-masking}, the mask is applied to the feature maps before the temporal spiking process, ensuring that masked regions produce no spike activity.

\textbf{Encoding.} The input video clip is first masked by the initial tube mask, $\boldsymbol{M}^0\cdot \boldsymbol{I}$, and then fed into the spike-driven video encoder. In the $l$-th intermediate block, the outputs from all basic layers are further modulated by the corresponding layer-wise tube mask $\boldsymbol{M}^l$. Therefore, the multi-scale latent features extracted by the spike-driven encoder are represented as $[\boldsymbol{U}^{1}, ..., \boldsymbol{U}^{4}]$. The process can be summarized as
\begin{equation}
    [\boldsymbol{U}^{1}, ..., \boldsymbol{U}^{4}] = \text{SpikeEncoder}\Big(\boldsymbol{I}, [\boldsymbol{M}^0, ..., \boldsymbol{M}^3]\Big).
\end{equation}
Unlike ViT-based encoders that convert frames into token sequences~\cite{tong2022videomae,wang2023videomae}, CNNs preserve spatial relationships during feature encoding, enabling stronger representation learning in the masked pretraining setting.


\textbf{Decoding.} Since our goal is to pretrain a spike-driven video encoder for downstream tasks, we adopt a standard ViT~\cite{dosovitskiy2020image} as the decoder for simplicity. During pretraining, masked regions in the encoder's output are filled with a learnable mask embedding $\boldsymbol{e}^{\text{mask}}$, after which the combined features are flattened, augmented with positional embeddings, and fed into the ViT decoder. The decoder then reconstructs the video clip $\boldsymbol{\hat{I}}$ following the unflattening operation, formulated as
\begin{equation}
\boldsymbol{\hat{I}} = \text{ViT-Decoder}\Big(\boldsymbol{M}^4 \cdot \boldsymbol{U}^{4} + (\boldsymbol{1} - \boldsymbol{M}^4) \cdot \boldsymbol{e}^{\text{mask}}\Big).
\end{equation}
\textbf{\textit{After pretraining, the ViT decoder is discarded}}, and only the spike-driven encoder is retained for downstream applications.

\revision{
The reconstruct loss is the mean squared error (MSE) over these masked regions, formulated as 
\begin{equation}
    \mathcal{L}_{\text{rec}} = \frac{\| (1-\boldsymbol{M}^0)\cdot(\boldsymbol{I} - \hat{\boldsymbol{I}}) \|_2^2}{\|1-\boldsymbol{M}^0\|^2_2}.
\end{equation}}



\revision{
\textbf{Multi-Spectral Knowledge Distillation.} To further enhance visual representation of SNN models in surgical scenes, we leverage the strong semantic encoding capability of SurgicalSAM2~\cite{liu2024surgical} via knowledge distillation. A key challenge lies in the mismatch between ANN and SNN representations, which hinders direct knowledge ill-posed. In contrast, semantic patterns are more stable in the frequency domain, where spatiotemporal variations are captured as spectral components. Motivated by this observation, we propose to transfer knowledge from the teacher ANN to the student SNN by aligning their representations across multiple spectral bands.}

\revision{
Given the intermediate feature maps of the teacher ANN and the student SNN, denoted as $\boldsymbol{U}^{\text{sam}}$ and $\boldsymbol{U}^4$, we first transform them into the spatial frequency domain using the Fast Fourier Transform (FFT). The resulting spectra are then partitioned into $B$ radial frequency bands by \textit{uniformly} dividing the spatial frequency range. In practice, we set $B=3$, corresponding to coarse structural patterns (low frequency), object-level semantics (mid frequency), and fine details (high frequency). The spectral components in each band are denoted as $\hat{\boldsymbol{U}}^{\text{sam}}_{b}$ and $\hat{\boldsymbol{U}}^4_{b}$, where $b$ indexes the $b$-th frequency band. Then we align the teacher and student representations within each band using the following distillation loss:
\begin{equation}
    \mathcal{L}_{\text{kd}}=\sum_{b=1}^{B} w_b \cdot \|\hat{\boldsymbol{U}}_{b}^{\text{sam}}-\hat{\boldsymbol{U}}_{b}^4\|_2^2,
\end{equation}
The band-adaptive weight $w_b=\|\hat{\boldsymbol{U}}_{b}^{\text{sam}}\|_2 / \sum_{i=1}^{B}\|\hat{\boldsymbol{U}}_{i}^{\text{sam}}\|_2$ is derived from the spectral energy of the teacher features to reflect the relative importance of different frequency ranges.}

\revision{
\textbf{Training.} The loss for pretraining consists of reconstruction loss and multi-spectral knowledge distillation loss, defined as
\begin{equation}
    \mathcal{L}_{\text{pretrain}} = \mathcal{L}_{\text{rec}} + \lambda\cdot \mathcal{L}_{\text{kd}},
\end{equation}
where $\lambda$ balances the contribution of the distillation term.}

\beforesubsection
\subsection{Surgical Video Segmentation Finetuning}
\aftersubsection
\label{sec:finetunning}

\revision{After spike-informed pretraining on surgical videos, the SNN encoder acquires more robust spatiotemporal representations for surgical scenes. Building on this pretrained encoder, we integrate a lightweight spike-driven segmentation head, consisting of a spike pyramid and a memory readout module, and then finetune the fully spike-driven model end-to-end for downstream surgical scene segmentation.}

\textbf{Video Segmentation Head.} To highlight the real-time potential of our SNN model for surgical scene segmentation, we avoid overly complex architectures, focusing instead on validating the effectiveness and efficiency of proposed approach. The segmentation head comprises two main components: 1) Memory Read and Fusion Module: Inspired by~\cite{paul2021local}, semantic features from previous frames (memory) are retrieved and fused with the current frame's features using spike-driven Hamming attention as detailed in Supplementary A, enabling temporal dependency modeling while preserving the event-driven efficiency of SNN. 2) Spike-Driven Feature Pyramid Networks (SpikeFPN): To maintain computational efficiency, we adapt the standard FPN~\cite{zhao2017pyramid} into a spike-driven design. 



\revision{
\textbf{Training.} To finetune the model for surgical video segmentation, we employ a combination of cross-entropy loss and focal loss to mitigate the severe class imbalance commonly observed in surgical datasets. Let $\boldsymbol{y}$ denote the one-hot representation of the ground-truth semantic label map. The losses are then defined as
\begin{gather}
\mathcal{L}_{\text{CE}} = - \frac{1}{THW}\sum_{t,i,j}^{H,W,T}\sum_{k=0}^{K-1} \boldsymbol{y}_{tijk}\cdot \log(\boldsymbol{\hat y}_{tijk}),  \\
\mathcal{L}_{\text{Focal}} = - \frac{1}{THW} \sum_{t,i,j}^{H,W,T} \sum_{k=0}^{K-1} \boldsymbol{y}_{tijk}\cdot (1 - \boldsymbol{\hat{y}}_{tijk})^\gamma \cdot \log(\boldsymbol{\hat{y}}_{tijk}),
\end{gather}
where $K$ denotes the number of semantic classes, $\boldsymbol{\hat{y}}_{tijk}$ is the predicted probability for class $k$ at pixel $(i,j)$ and time $t$, and $\gamma=2$ is the focusing parameter. The cross-entropy loss $\mathcal{L}_{\text{CE}}$ enforces alignment between predicted and ground-truth distributions, while the focal loss $\mathcal{L}_{\text{Focal}}$ down-weights easy examples and emphasizes hard pixels, thereby improving performance on under-represented classes.}

\beforesection
\section{Experiments}
\aftersection
\label{sec:exp}

\beforesubsection
\subsection{Experimental Details}\label{sec:exp-details}
\aftersubsection

\textbf{Datasets.} We evaluate our SNN framework on a pubilic avalable dataset for fair comparison with baselines and a private in-house dataset to illustrate the effectiveness of real-world application. Details are provided as follows:

\textit{1) EndoVis18 Dataset.} The 2018 MICCAI EndoVis Scene Segmentation Challenge (EndoVis18) dataset~\cite{allan20202018} features more complex surgical scenes and has been widely adopted in prior studies~\cite{liu2024surgical,yue2024surgicalsam}. It comprises 15 videos, each containing 149 frames of resolution $1024\times 1280$. Following the standard protocol in~\cite{gonzalez2020isinet}, we use sequences 2, 5, 9, and 15 for testing, while the remaining sequences are used for training. The instrument categories evaluated include Bipolar Forceps (BF), Prograsp Forceps (PF), Large Needle Driver (LND), Suction Instrument (SI), Clip Applier (CA), Monopolar Curved Scissors (MCS), and Ultrasound Probe (UP). 

\textit{2) In-House SurgBleed Dataset.} We curated an in-house dataset of hepatobiliary surgical videos, named the Surgical Bleeding Tracking (SurgBleed) dataset, to facilitate bleeding area segmentation and surgical alert applications. The dataset comprises 70 videos collected from 25 surgical cases, recorded at 30 frames per second with a resolution of $720\times 1280$ pixels and an overall duration of 4,721.6 seconds. Expert surgeons manually annotated the bleeding regions using polygonal masks every 10 frames, resulting in 6,703 labeled frames. Frames without bleeding were excluded from the annotated clips. On average, the annotated bleeding regions occupy 35,012.86 pixels per frame, corresponding to roughly a $187\times187$ pixel area or 3.79\% of the full frame. As representative examples illustrated in Supplementary B, the main challenges arise from the irregular and highly dynamic bleeding boundaries, \ie, large variations in bleeding size and frequent occlusions from surgical instruments. \textit{This in-house dataset is designed to better reflect real-world surgical environments}, where accurate and timely small bleeding area segmentation is far more critical than instrument recognition, as it directly impacts surgical decision-making and patient safety. In the experiment, a patient-level data split is applied, with approximately 80\% of the cases used for training and the remaining 20\% for testing. This yields 4,753 qualified 4-frame clips for training and 910 clips for testing. 

\textbf{Implementation Details.} Our model is partially implemented using SpikingJelly~\cite{SpikingJelly}. We employ a parametric LIF neuron with a soft reset, where backpropagation through the reset path is preserved. In the segmentation head, the SpikeFPN module utilizes an IntLIF neuron with $Z=4$. During training, we apply several data augmentation strategies, including random resizing within the range of 0.3–0.7, cropping to an input size of $512\times640$ for the EndoVis18 dataset and $384\times640$ for the SurgBleed dataset, random horizontal flipping, and photometric distortion. Each input video clip contains 4 frames (\ie, $T=4$), consistent with prior works on video semantic segmentation~\cite{gonzalez2020isinet,zou2025spikevideoformer,ravi2024sam2}. 


\revision{
For spike-informed MAE pretraining, \textbf{\textit{we use only the original training set without introducing additional unlabeled data}}. This setup ensures a fair comparison with models trained directly on the surgical video segmentation task, as our primary goal is to demonstrate the effectiveness of video-masked pretraining within the SNN framework. During pretraining, the temperature $\tau$ for masking region selection is initialized at 3 and linearly annealed to 0.3 across epochs, gradually shifting from smooth to more selective masking. The knowledge distillation weight $\lambda$ is set to 0.1. We train the model on 4 or 8 NVIDIA A6000 GPUs with a batch size of 16 and a learning rate of $4\times10^{-4}$ for the 16M and 56.3M spike-driven encoder variants, respectively. The decoder is an 8-layer ViT~\cite{dosovitskiy2020image} with space–time self-attention for video reconstruction. Pretraining runs for 200 epochs under a cosine annealing schedule (maximum 250 epochs).}

For surgical video segmentation finetuning, the memory read and fusion module outputs 128 or 256 channels, which are subsequently processed by the SpikeFPN with matching channel dimensions for the 16M and 56.3M encoder variants, respectively. The final layer produces per-pixel class probabilities, \ie, 8 classes for the EndoVis18 dataset and 2 classes for the SurgBleed dataset. Training on both datasets is conducted with a batch size of 16 and a learning rate of $4\times10^{-4}$, using 4 or 8 NVIDIA A6000 GPUs for the two model variants. The models are finetuned for 80 epochs under a cosine annealing schedule with a maximum of 120 epochs.

\textbf{Evaluation Metrics.} We evaluate the segmentation performance using the Intersection over Union (\textbf{IoU}), which quantifies the overlap between predicted and ground-truth regions as the ratio of their intersection to their union. For the EndoVis18 dataset, both class-wise IoU and mean IoU (\textbf{mIoU}) are reported to provide a comprehensive assessment. To evaluate the efficiency of SNN, we report power consumption (\textbf{Power}) in milli-joules (mJ), estimated based on the number of AC or MAC floating-point operations required by each model. Inference time (\textbf{Latency}) in milli-seconds (ms) is also reported to assess runtime performance, estimated on the AMD Xilinx ZCU104 platform~\cite{li2023firefly} as is described in Sec.~\ref{sec:snns}. This setup excludes GPU support, ensuring a fair evaluation of computational efficiency for real-world applications of SNN. 

\begin{table*}[!ht]
    \centering
    \caption{\revision{\textbf{Surgical Instrument Segmentation on EndoVis18 Dataset.} Our SNN models achieve comparable mIoU with most ANN models while significantly reducing inference time and power consumption. For fairness, all prompt-based baselines use a single-point prompt as additional input. Note that TrackAnything~\cite{yang2023track}, PerSAM~\cite{zhang2023personalize} and SurgicalSAM~\cite{yue2024surgicalsam} uses SAM-B as the base model, whereas SurgicalSAM2~\cite{liu2024surgical} adopts the most lightweight SAM2-Small as its base model.}
    }
    \vspace{-2mm}
    \setlength{\tabcolsep}{1mm}
    \renewcommand{\arraystretch}{1.2}
    \resizebox{\textwidth}{!}{
    \begin{tabular}{@{}|c|c|c|ccccc|ccccccc|@{}}
    \bottomrule \hline
        \multirow{2}{*}{\makecell[c]{Models}} & \multirow{2}{*}{\makecell[c]{Methods}} & \multirow{2}{*}{\makecell[c]{SNN}} & \multirow{2}{*}{\makecell[c]{Params}} & \multirow{2}{*}{\makecell[c]{Power\\(mJ)}} & \multirow{2}{*}{\makecell[c]{Latency\\(ms)}} & \multirow{2}{*}{\makecell[c]{cIoU $\uparrow$}} & \multirow{2}{*}{\makecell[c]{mIoU $\uparrow$}}  & \multicolumn{7}{c|}{Instrument Category IoU $\uparrow$}\\
    \cline{9-15}
        & & & & & & & & \makecell[c]{BF} & \makecell[c]{PF} & \makecell[c]{LND} & \makecell[c]{SI} & \makecell[c]{CA} & \makecell[c]{MCS} & \makecell[c]{UP}\\
    \hline
        \multirow{4}{*}{\makecell[c]{Prompt-\\Based\\Models}} & \makecell[c]{TrackAnything~\cite{yang2023track}} & \makecell[c]{\xmark} & \makecell[c]{91.0M} & \makecell[c]{533.3} & \makecell[c]{167.7} & \makecell[c]{40.36} & \makecell[c]{20.62} & \makecell[c]{30.20} & \makecell[c]{12.87} & \makecell[c]{24.46} & \makecell[c]{9.17} & \makecell[c]{0.19} & \makecell[c]{55.03} & \makecell[c]{12.41} \\ 
        & \makecell[c]{PerSAM~\cite{zhang2023personalize}} & \makecell[c]{\xmark} & \makecell[c]{91.0M} & \makecell[c]{533.3} & \makecell[c]{167.7} & \makecell[c]{49.21} & \makecell[c]{34.55} & \makecell[c]{51.26} & \makecell[c]{34.40} & \makecell[c]{\textbf{46.75}} & \makecell[c]{16.45} & \makecell[c]{\textbf{15.07}} & \makecell[c]{52.28} & \makecell[c]{\textbf{25.62}} \\ 
        & \makecell[c]{SurgicalSAM~\cite{yue2024surgicalsam}} & \makecell[c]{\xmark} & \makecell[c]{93.5M} & \makecell[c]{560.6} & \makecell[c]{176.3} & \makecell[c]{74.12} & \makecell[c]{43.67} & \makecell[c]{80.32} & \makecell[c]{50.78} & \makecell[c]{19.73} & \makecell[c]{43.81} & \makecell[c]{5.23} & \makecell[c]{88.56} & \makecell[c]{17.32} \\ 
        & \makecell[c]{SurgicalSAM2~\cite{liu2024surgical}} & \makecell[c]{\xmark} & \makecell[c]{49.3M} & \makecell[c]{208.4} & \makecell[c]{65.6} & \makecell[c]{70.60} & \makecell[c]{37.69} & \makecell[c]{77.23} & \makecell[c]{\textbf{52.63}} & \makecell[c]{0.12} & \makecell[c]{42.78} & \makecell[c]{0.00} & \makecell[c]{91.07} & \makecell[c]{0.00} \\ 
    \hline
        \multirow{5}{*}{\makecell[c]{Task-\\Specific\\Models}} & \makecell[c]{ISINet~\cite{gonzalez2020isinet}} & \makecell[c]{\xmark} & \makecell[c]{162.5M} & \makecell[c]{1527.2} & \makecell[c]{480.3} & \makecell[c]{73.03} & \makecell[c]{40.21} & \makecell[c]{73.83} & \makecell[c]{48.61} & \makecell[c]{30.98} & \makecell[c]{37.68} & \makecell[c]{0.00} & \makecell[c]{88.16} & \makecell[c]{2.16} \\
        & \makecell[c]{S3Net~\cite{baby2023forks}} & \makecell[c]{\xmark} & \makecell[c]{71.4M} & \makecell[c]{524.4} & \makecell[c]{164.9} & \makecell[c]{75.81} & \makecell[c]{42.58} & \makecell[c]{77.22} & \makecell[c]{50.87} & \makecell[c]{19.83} & \makecell[c]{50.59} & \makecell[c]{0.00} & \makecell[c]{92.12} & \makecell[c]{7.44} \\
        & \makecell[c]{MATIS-Frame~\cite{ayobi2023matis}} & \makecell[c]{\xmark} & \makecell[c]{97.0M} & \makecell[c]{588.8} & \makecell[c]{185.2} & \makecell[c]{\textbf{82.37}} & \makecell[c]{\textbf{48.65}} & \makecell[c]{\textbf{83.35}} & \makecell[c]{38.82} & \makecell[c]{40.19} & \makecell[c]{\textbf{64.49}} & \makecell[c]{4.32} & \makecell[c]{\textbf{93.18}} & \makecell[c]{16.17} \\
    \cline{2-15}
        & \multirow{2}{*}{\makecell[c]{SpikeSurgSeg\\(Ours)}} & \makecell[c]{\cmark} & \makecell[c]{\textbf{21.1M}} & \makecell[c]{\textbf{40.8}} & \makecell[c]{\textbf{8.2}} & \makecell[c]{74.56} & \makecell[c]{42.11} & \makecell[c]{81.87} & \makecell[c]{47.32} & \makecell[c]{40.31} & \makecell[c]{30.81} & \makecell[c]{1.74} & \makecell[c]{91.70} & \makecell[c]{1.0} \\ 
        & & \makecell[c]{\cmark} & \makecell[c]{62.1M} & \makecell[c]{178.8} & \makecell[c]{35.9} & \makecell[c]{77.32} & \makecell[c]{43.47} & \makecell[c]{83.30} & \makecell[c]{48.53} & \makecell[c]{41.95} & \makecell[c]{30.66} & \makecell[c]{5.38} & \makecell[c]{92.30} & \makecell[c]{2.15} \\
    \hline \toprule 
    \end{tabular}}
    \label{tab:endovis18}
    \vspace{-4mm}
\end{table*}

\beforesubsection
\subsection{Comparison on EndoVis18 Dataset}\label{sec:compare-endovis}
\aftersubsection

\textbf{Baselines.} Our SpikeSurgSeg is a task-specific model. Accordingly, we compare it with several task-specific ANN baselines, including ISINet~\cite{gonzalez2020isinet}, S3Net~\cite{baby2023forks}, and MATIS~\cite{ayobi2023matis}. In contrast, recent methods have mostly shifted toward prompt-based paradigms that adapt SAM or SAM2 for downstream tasks and rely on user-provided prompts as auxiliary inputs. We therefore include TrackAnything~\cite{yang2023track}, PerSAM~\cite{zhang2023personalize}, SurgicalSAM~\cite{yue2024surgicalsam}, and SurgicalSAM2~\cite{liu2024surgical} as additional baselines. Since prompts convey explicit prior knowledge about target objects, these models generally yield higher accuracy. For a fair comparison, we adopt the simplest configuration, \ie, a single-point prompt, for all prompt-based baselines.

\begin{figure}[!t]
    \centering
    \includegraphics[width=\columnwidth]{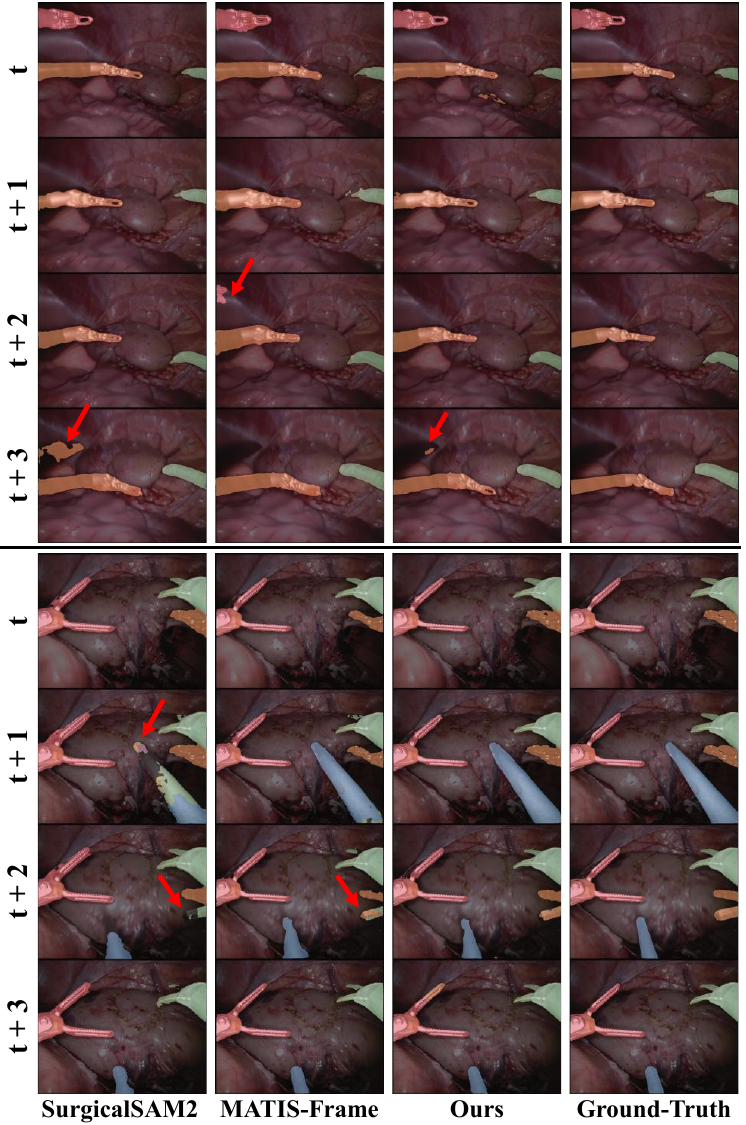}
    \vspace{-4mm}
    \caption{\textbf{Qualitative Results on EndoVis18 Dataset.} Our SNN model maintains high efficiency while reducing false positives and inference time, benefiting from the inherent sparsity of spike-driven computation.
    }
    \vspace{-5mm}
    \label{fig:results-endovis}
\end{figure}

\revision{
\textbf{Results.} As shown in Table~\ref{tab:endovis18}, SpikeSurgSeg achieves a markedly more balanced trade-off between accuracy, latency, and energy efficiency compared with both prompt-based and task-specific ANN baselines. Prompt-based SAM variants exhibit strong performance on certain categories due to prompt-induced priors (\eg, PerSAM achieves 34.55\% mIoU), but their results are inconsistent across instruments and incur substantial computational overhead (\eg, $>$500 mJ power and $>$160 ms latency). Although SurgicalSAM2 reduces computation (208.4 mJ, 65.6 ms), its overall accuracy remains limited (37.69\% mIoU) and degrades significantly on LND and UP categories, suggesting that lightweight foundation models struggle with fine-grained surgical segmentation. Task-specific ANN models (ISINet, S3Net, and MATIS) achieve higher accuracy (up to 48.65\% mIoU), but at the cost of heavy computation (524–1527 mJ, 165–480 ms), limiting their practicality for real-time deployment. In contrast, the small SpikeSurgSeg model attains competitive performance (42.11\% mIoU) with only 21.1M parameters, while reducing power consumption by 13–37$\times$ and latency by 20–58$\times$ compared to ANN counterparts. The larger variant further improves accuracy (43.47\% mIoU) and category-level IoUs, while maintaining low power (178.8 mJ) and latency (35.9 ms), satisfying real-time constraints. These results demonstrate that spike-driven models can approach the accuracy of SOTA methods while offering significantly improved efficiency, making them well-suited for deployable surgical applications.
}

\begin{figure*}[!t]
    \centering
    \includegraphics[width=\textwidth]{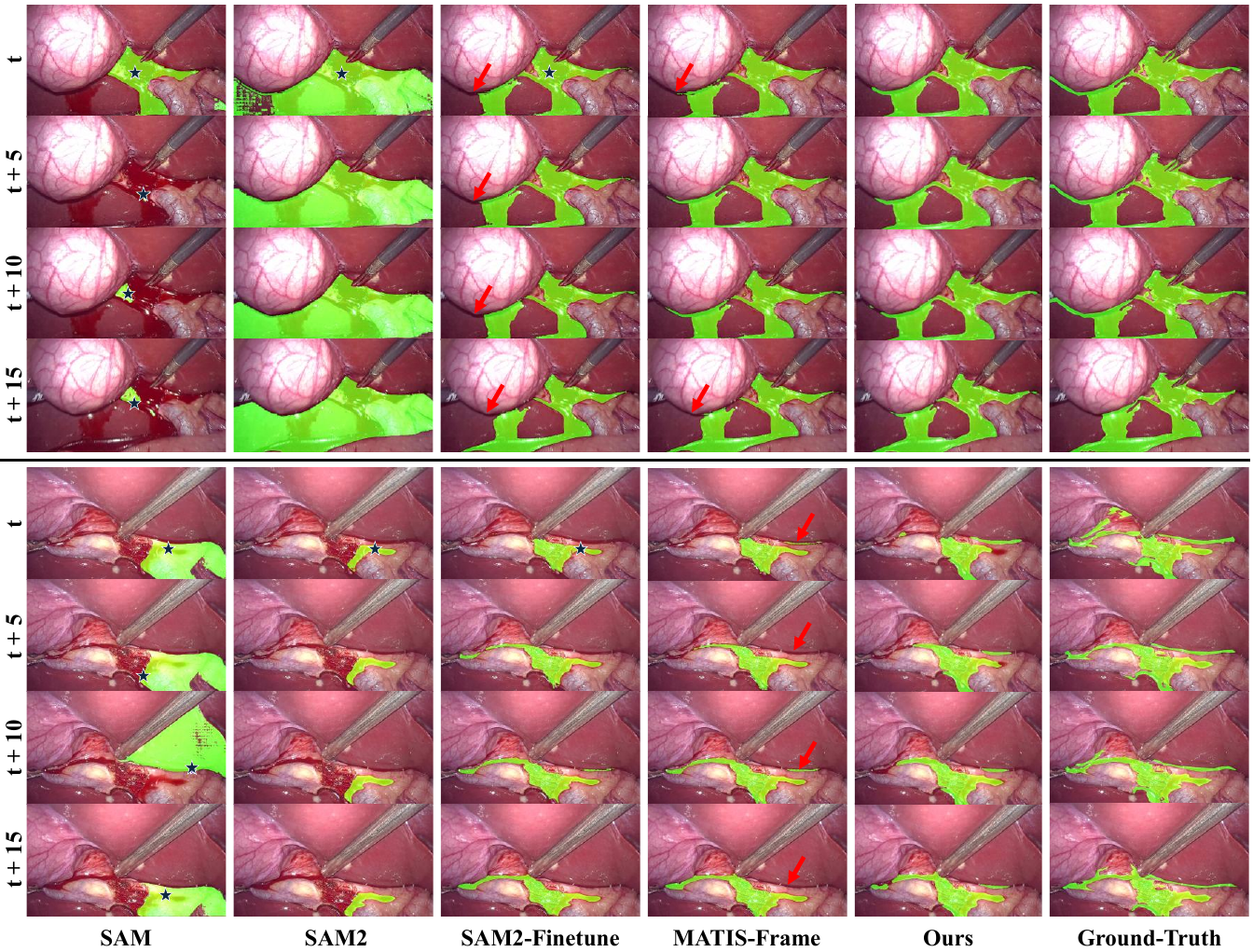}
    \caption{\textbf{Qualitative Results on SurgBleed Dataset.} Compared with the task-specific MATIS-Frame method and the prompt-based SAM2-Finetune method, our SNN model achieves comparable or even superior bleeding segmentation performance, while maintaining substantially lower energy consumption and inference latency.
    }
    \vspace{-4mm}
    \label{fig:results-bleeding}
\end{figure*}

\begin{table}[!t]
    \centering
    \caption{\revision{\textbf{Surgical Bleeding Segmentation on SurgBleed Dataset.} We use SAM-Base model for the SAM baseline abd SAM2-Base+ for all SAM2 baselines.}
    }
    \vspace{-2mm}
    \setlength{\tabcolsep}{0.5mm}
    \renewcommand{\arraystretch}{1.2}
    \resizebox{\columnwidth}{!}{
    \begin{tabular}{@{}|c|c|c|cccc|@{}}
    \bottomrule \hline
        \multirow{2}{*}{\makecell[c]{Models}} & \multirow{2}{*}{\makecell[c]{Methods}} & \multirow{2}{*}{\makecell[c]{SNN}} & \multirow{2}{*}{\makecell[c]{Params}} & \multirow{2}{*}{\makecell[c]{Power\\(mJ)}} & \multirow{2}{*}{\makecell[c]{Latency\\(ms)}} & \multirow{2}{*}{\makecell[c]{mIoU $\uparrow$}} \\
        & & & & & & \\
    \hline
        \multirow{4}{*}{\makecell[c]{Prompt-\\Based\\Models}} & \makecell[c]{SAM~\cite{kirillov2023segment}} & \makecell[c]{\xmark} & \makecell[c]{91.0M} & \makecell[c]{400.0} & \makecell[c]{125.8} & \makecell[c]{28.40} \\         
        & \makecell[c]{SAM2~\cite{ravi2024sam2}} & \makecell[c]{\xmark} & \makecell[c]{80.8M} & \makecell[c]{293.3} & \makecell[c]{92.2} & \makecell[c]{32.10} \\ 
        & \makecell[c]{SAM2-Finetune~\cite{liu2024surgical}} & \makecell[c]{\xmark} & \makecell[c]{80.8M} & \makecell[c]{293.3} & \makecell[c]{92.2} & \makecell[c]{69.79} \\ 
        & \makecell[c]{SAM2-Adapter~\cite{chen2023sam}} & \makecell[c]{\xmark} & \makecell[c]{81.3M} & \makecell[c]{308.1} & \makecell[c]{96.9} & \makecell[c]{69.40} \\ 
    \hline
        \multirow{4}{*}{\makecell[c]{Task-\\Specific\\Models}} & \makecell[c]{ISINet~\cite{gonzalez2020isinet}} & \makecell[c]{\xmark} & \makecell[c]{162.5M} & \makecell[c]{1145.4} & \makecell[c]{360.2} & \makecell[c]{64.95} \\
        & \makecell[c]{MATIS-Frame~\cite{ayobi2023matis}} & \makecell[c]{\xmark} & \makecell[c]{97.0M} & \makecell[c]{441.6} & \makecell[c]{138.9} & \makecell[c]{70.80} \\
    \cline{2-7}
        & \multirow{2}{*}{\makecell[c]{SpikeSurgSeg\\(Ours)}} & \makecell[c]{\cmark} & \makecell[c]{\textbf{21.1M}} & \makecell[c]{\textbf{30.6}} & \makecell[c]{\textbf{6.1}} & \makecell[c]{69.70} \\ 
        & & \makecell[c]{\cmark} & \makecell[c]{62.1M} & \makecell[c]{116.1} & \makecell[c]{23.3} & \makecell[c]{\textbf{70.85}} \\
    \hline \toprule 
    \end{tabular}}
    \vspace{-5mm}
    \label{tab:surgbleed}
\end{table}

\beforesubsection
\subsection{Comparison on SurgBleed Dataset}\label{sec:compare-surgbleed}
\aftersubsection

\textbf{Baselines.} For prompt-based models, we evaluate two types of SAM-based results. The first is zero-shot inference, in which SAM-Base~\cite{kirillov2023segment} and SAM2-Base+~\cite{ravi2024sam2} are directly tested on the SurgBleed dataset using a single-point prompt derived from the ground-truth bleeding mask. The second is finetuned performance, where SAM2-Base+ is trained on the SurgBleed dataset for 100 epochs with a learning rate of $5\times10^{-5}$ following SurgicalSAM2~\cite{liu2024surgical}. Additionally, we also include finetuning with the adapter proposed in~\cite{chen2023sam}. For task-specific models, ISINet~\cite{gonzalez2020isinet} and MATIS~\cite{ayobi2023matis} are re-trained on the SurgBleed dataset.


\revision{
\textbf{Results.} As shown in Tab.~\ref{tab:surgbleed}, SpikeSurgSeg achieves competitive performance while offering substantial efficiency gains. The lightweight 21.1M variant reaches 69.70\% mIoU, only slightly below the task-specific SOTA MATIS (70.80\%), yet reduces power consumption and latency by approximately 14.4$\times$ and 22.8$\times$, respectively. The larger 62.1M variant further improves performance to 70.85\% mIoU, surpassing MATIS while still maintaining significantly lower computational cost (3.8$\times$ lower power and 6.0$\times$ faster inference). In the prompt-based setting, SAM (28.40\%) lags behind SAM2 (32.10\%) by 3.7\% mIoU, highlighting the importance of video-aware modeling for surgical understanding. After fine-tuning, SAM2 achieves strong performance (69.79\%), but at the expense of high computation (293.3 mJ, 92.2 ms). In contrast, our SNN model attains comparable accuracy while reducing power consumption by over $9.6\times$ and latency by $15.1\times$, even without relying on additional user prompts.}

Qualitative results in Fig.~\ref{fig:results-bleeding} further demonstrate that our method produces accurate and temporally stable segmentation. Overall, these results highlight the advantage of spike-driven architectures in achieving near-SOTA performance with drastically improved efficiency, making them well-suited for real-time surgical deployment.

\beforesubsection
\subsection{Ablation Study}\label{sec:ablation}
\aftersubsection



\begin{figure*}[!t]
    \centering
    \includegraphics[width=0.9\textwidth]{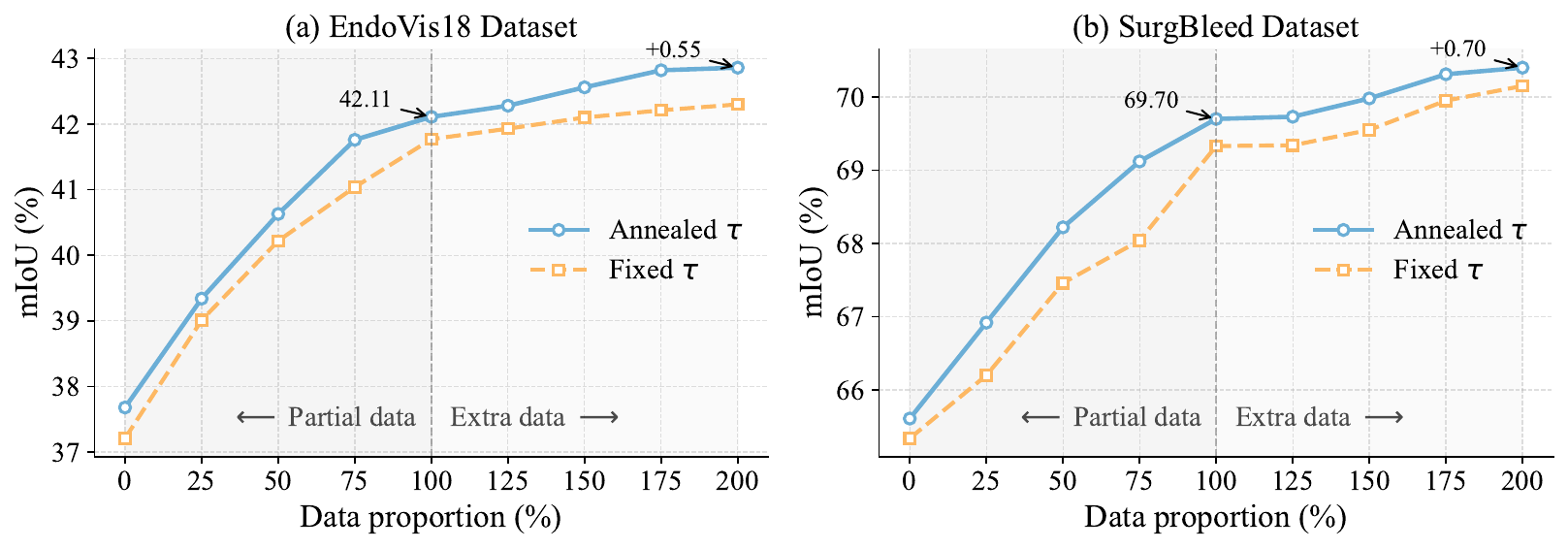}
    \vspace{-2mm}
    \caption{\revision{\textbf{Ablation of Pretraining Data Size.} We report downstream surgical segmentation performance (mIoU) as the pretraining data size varies from 0–200\% on (a) EndoVis18 and (b) SurgBleed, under two spike-informed masking strategies (annealing $\tau=3\rightarrow0.3$ and fixed $\tau=1$). 100\% porportion data means the original training set. Extra data is drawn from EndoVis17 for (a) and unlabeled frames from our in-house dataset for (b). In the left panels (Partial data) of each sub-figure, increasing pretraining data consistently improves mIoU, whereas in the right panels (Extra data), the gains become less pronounced. Overall, these results highlight the value of large-scale unlabeled surgical video data for enhancing spatiotemporal representation learning in SNN models, thereby narrowing the performance gap with ANN while preserving low energy consumption and latency.}
    }
    \vspace{-3mm}
    \label{fig:results-ablation}
\end{figure*}

\begin{figure}[!t]
    \centering
    \includegraphics[width=\columnwidth]{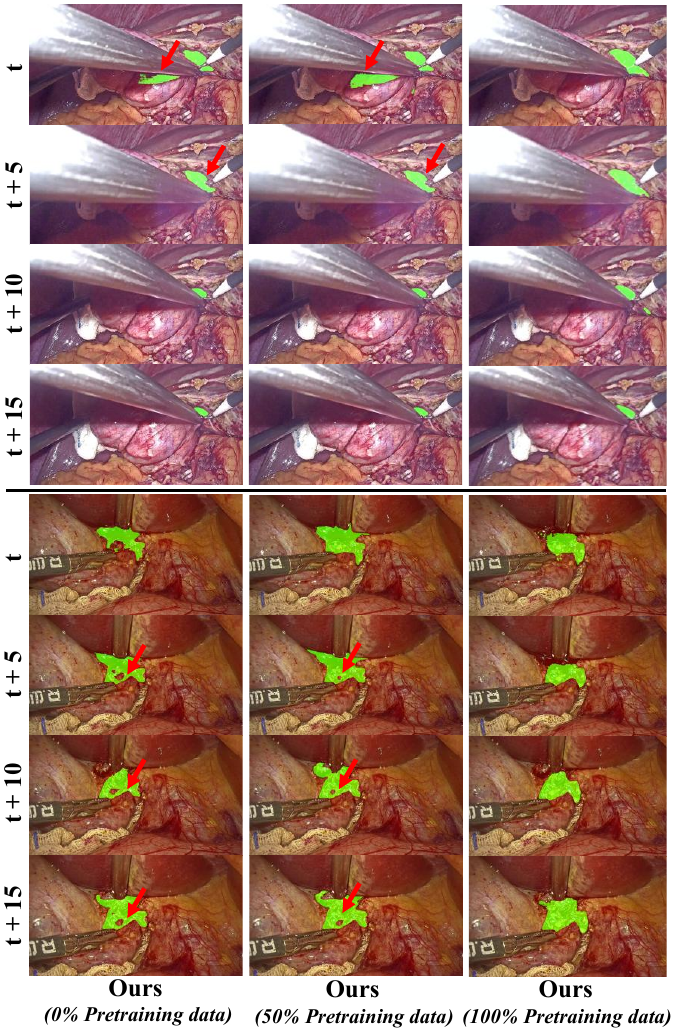}
    \vspace{-5mm}
    \caption{\textbf{Qualitative Comparison with Varying Pretraining Data.} Using 0\% or only 50\% of the data for pretraining results in fragmented and temporally inconsistent predictions, whereas pretraining on the full dataset (100\%) improves the accuracy and coherence of segmentation results.
    }
    \vspace{-4mm}
    \label{fig:qualitative-ablation}
\end{figure}

\subsubsection{Pretraining Stage} \revision{To better understand what drives effective representation learning in our spike-informed pretraining framework, we perform a series of ablations. We analyze how pretraining data scale, unlabeled video augmentation, mask ratio, and KD loss each contribute to downstream segmentation accuracy as follows.}
\begin{itemize}
    \item \revision{\textbf{Partial Pretraining Data.} To evaluate the impact of pretraining data size on representation quality, we pretrain the model using only subsets of the training data. As shown in Fig.~\ref{fig:results-ablation} (a) and (b) \textit{left panels}, increasing the pretraining data consistently improves mIoU on both EndoVis18 and SurgBleed under annealing and fixed masking strategies. This trend indicates that large-scale video exposure is critical for learning robust spatiotemporal representations. Qualitative results in Fig.~\ref{fig:qualitative-ablation} further support this observation, showing that limited pretraining data often leads to fragmented boundaries and temporally inconsistent predictions.}
    
    
    \item \revision{\textbf{Extra Pretraining Data.} We further examine the impact of incorporating extra unlabeled data during pretraining. Specifically, for EndoVis18, we include videos from EndoVis17~\cite{allan20202018}, while for SurgBleed, we use unlabeled frames from our in-house SurgBleed dataset. The amount of extra data is controlled from 100\% to 200\% relative to the original training set. As shown in Fig.~\ref{fig:results-ablation} (a) and (b) \textit{right panels}, increasing the amount of unlabeled data yields consistent, albeit moderate, performance improvements on both datasets. These results highlight the potential of large-scale pretraining to enhance surgical scene understanding in SNN models, helping to narrow the performance gap with ANN while preserving low energy consumption and latency.}

    \item \revision{\textbf{Masking Strategy.} We compare three masking strategies in Tab.~\ref{tab:ablation-masking}: fully random masking as used in prior work~\cite{tong2022videomae,wang2023videomae}, and our spike-informed masking with fixed and annealed $\tau$. Spike-informed masking consistently outperforms random masking on both datasets, highlighting the benefit of leveraging spike activity to guide masking for SNN pretraining. Within the fixed setting, larger $\tau$ generally yields better results, while the annealing strategy achieves the best performance overall. This suggests that early training benefits from more uniform masking, whereas later stages gain from masking more informative regions, encouraging the SNN encoder to focus on more salient context. Fig.~\ref{fig:results-ablation} further confirms that annealed $\tau$ consistently outperforms fixed $\tau$ across different pretraining data sizes.}
    
    
    \item \revision{\textbf{Multi-Spectral KD Loss.} As shown in Tab.~\ref{tab:ablation-kd}, using KD ($B=1$) significantly improves both IS and mIoU compared to the baseline without KD. Further increasing the number of spectral bands to $B=3$ consistently yields the best performance across both datasets, indicating that band-wise alignment provides more fine-grained and effective knowledge transfer. Increasing $B$ to $5$ brings no additional gains, suggesting that a moderate number of frequency bands is sufficient to capture the essential spectral structure while avoiding unnecessary complexity.}
    
\end{itemize}

\begin{table}[!t]
    \centering
    \caption{\textbf{Ablation of Masking Strategy.} The best performance setting (in \textbf{bold}) is adopted in our method.}
    \vspace{-2mm}
    \setlength{\tabcolsep}{1.0mm}
    \renewcommand{\arraystretch}{1.2}
    \resizebox{\columnwidth}{!}{
    \begin{tabular}{@{}|c|c|c|@{}}
    \bottomrule \hline
        \multirow{2}{*}{\makecell[c]{Making Strategy}} & \makecell[c]{EndoVis18} & \makecell[c]{SurgBleed} \\
    \cline{2-3}
        & \makecell[c]{mIoU $\uparrow$} & \makecell[c]{mIoU $\uparrow$} \\
    \hline
        \makecell[l]{Fully Random (ratio 0.3)} & \makecell[c]{40.79} & \makecell[c]{68.21} \\
        \makecell[l]{Fully Random (ratio 0.5)} & \makecell[c]{41.65} & \makecell[c]{69.12} \\
    \hline
        \makecell[l]{Spike-Informed (fixed $\tau=0.3$)} & \makecell[c]{40.95} & \makecell[c]{69.01} \\ 
        \makecell[l]{Spike-Informed (fixed $\tau=1$)} & \makecell[c]{41.77} & \makecell[c]{69.33} \\ 
        \makecell[l]{Spike-Informed (fixed $\tau=3$)} & \makecell[c]{41.93} & \makecell[c]{69.40} \\ 
    \hline
        \makecell[l]{Spike-Informed (annealing $\tau=3\rightarrow1$)} & \makecell[c]{42.08} & \makecell[c]{69.61} \\
        \makecell[l]{Spike-Informed (annealing $\tau=3\rightarrow0.3$)} & \makecell[c]{\textbf{42.11}} & \makecell[c]{\textbf{69.70}} \\ 
        \makecell[l]{Spike-Informed (annealing $\tau=1\rightarrow0.3$)} & \makecell[c]{42.03} & \makecell[c]{69.48} \\ 
    \hline \toprule 
    \end{tabular}}
    \label{tab:ablation-masking}
    \vspace{-3mm}
\end{table}

\begin{table}[!t]
    \centering
    \caption{\textbf{Ablation of Multi-Spectral KD Loss.} IS denotes the Inception Score measuring video reconstruction fidelity.}
    \vspace{-2mm}
    \setlength{\tabcolsep}{2.5mm}
    \renewcommand{\arraystretch}{1.2}
    \resizebox{\columnwidth}{!}{
    \begin{tabular}{@{}|c|c|cc|cc|@{}}
    \bottomrule \hline
        \multirow{2}{*}{\makecell[c]{KD\\Loss}} & \multirow{2}{*}{\makecell[c]{Spectral\\Bands}} & \multicolumn{2}{c|}{\makecell[c]{EndoVis18}} & \multicolumn{2}{c|}{\makecell[c]{SurgBleed}} \\
    \cline{3-6}
        & & \makecell[c]{IS $\downarrow$} & \makecell[c]{mIoU $\uparrow$} & \makecell[c]{IS $\downarrow$} & \makecell[c]{mIoU $\uparrow$} \\
    \hline
        \makecell[c]{\xmark} & - & \makecell[c]{0.19} & \makecell[c]{40.62} & \makecell[c]{0.17} & \makecell[c]{68.57} \\
        \makecell[c]{\cmark} & $B=1$ & \makecell[c]{0.14} & \makecell[c]{42.04} & \makecell[c]{0.12} & \makecell[c]{69.43} \\ 
        \makecell[c]{\cmark} & $B=3$ & \makecell[c]{\textbf{0.13}} & \makecell[c]{\textbf{42.11}} & \makecell[c]{\textbf{0.12}} & \makecell[c]{\textbf{69.70}} \\ 
        \makecell[c]{\cmark} & $B=5$ & \makecell[c]{0.13} & \makecell[c]{42.11} & \makecell[c]{0.12} & \makecell[c]{69.70} \\ 
    \hline \toprule 
    \end{tabular}}
    \label{tab:ablation-kd}
    \vspace{-3mm}
\end{table}


\subsubsection{Finetunning Stage} We analyze the behavior of the pretrained SNN encoder on downstream surgical tasks, focusing on its generalization to other task and the effect of freezing versus updating the backbone during segmentation.
\begin{itemize}
    \item \textbf{Generalization to Other Downstream Task.} To assess how well the pretrained SNN encoder transfers beyond segmentation, we finetune it for two recognition tasks: \textit{7-instrument classification} on EndoVis18 and \textit{bleeding detection} on the extended SurgBleed dataset (balanced 1:1 for bleeding vs.\ non-bleeding). A spiking linear layer is attached to the encoder output, and performance is measured by (mean) accuracy. As shown in Tab.~\ref{tab:ablation-recognition}, pretraining yields clear gains on both tasks, improving instrument recognition from 73.5 to 81.2 and bleeding detection from 86.0 to 91.7. These consistent improvements demonstrate that the pretrained representations capture transferable spatiotemporal cues beneficial across diverse surgical video understanding tasks.
    
    \item \textbf{SNN Backbone Freezing.} We further investigate whether the pretrained SNN encoder should be frozen or updated during downstream fine-tuning. As shown in Tab.~\ref{tab:ablation-freeze}, updating the backbone yields slight mIoU improvements on both EndoVis18 and SurgBleed, while the frozen variant achieves acceptable performance. This suggests that pretraining provides stable and high-quality representations of surgical scenes.
\end{itemize}

\beforesubsection
\subsection{Discussion}
\aftersubsection
\label{sec:discussion}

\revision{
\textbf{Computational Redundancy.} Although recent ANN foundation models demonstrate strong segmentation performance, our experiments show that their large-scale architectures incur substantial computational and energy overhead for surgical video segmentation. As shown in Tab.~\ref{tab:endovis18} and~\ref{tab:surgbleed}, this overhead makes it difficult to satisfy the strict latency and power constraints of resource-constrained operative environments. In contrast, \textit{SpikeSurgSeg} achieves comparable segmentation accuracy with much lower latency and energy consumption through a fully spike-driven design and tailored spike-informed pretraining. These results suggest that, for deployable surgical systems, efficiency-oriented neuromorphic architectures may offer a more practical solution than scaling up conventional foundation models.}

\begin{table}[!t]
    \centering
    \caption{\revision{\textbf{Pretraining Generalization to Downstream Classification Task.} We evaluate the effect of pretraining on recognition by comparing an SNN backbone with and without pretraining after finetuning for 7-instrument classification on EndoVis18 and bleeding recognition on SurgBleed.}
    }
    \vspace{-2mm}
    \setlength{\tabcolsep}{1.5mm}
    \renewcommand{\arraystretch}{1.2}
    \resizebox{\columnwidth}{!}{
    \begin{tabular}{@{}|c|c|c|@{}}
    \bottomrule \hline
        \multirow{2}{*}{\makecell[c]{Use\\Pretraining}} & \makecell[c]{Instrument (EndoVis18)} & \makecell[c]{Bleeding (SurgBleed)} \\
    \cline{2-3}
        & \makecell[c]{Accuracy $\uparrow$} & \makecell[c]{Accuracy $\uparrow$} \\
    \hline
        \makecell[c]{\xmark} & \makecell[c]{81.2} & \makecell[c]{91.7} \\ 
        \makecell[c]{\cmark} & \makecell[c]{73.5} & \makecell[c]{86.0} \\
    \hline \toprule 
    \end{tabular}}
    \vspace{-2mm}
    \label{tab:ablation-recognition}
\end{table}

\begin{table}[!t]
    \centering
    \caption{\textbf{Ablation of SNN Encoder Freezing.} We compare freezing and unfreezing encoder during finetuning.}
    \vspace{-2mm}
    \setlength{\tabcolsep}{6mm}
    \renewcommand{\arraystretch}{1.2}
    \resizebox{\columnwidth}{!}{
    \begin{tabular}{@{}|c|c|c|@{}}
    \bottomrule \hline
        \multirow{2}{*}{\makecell[c]{SNN Encoder\\Freezing}} & \makecell[c]{EndoVis18} & \makecell[c]{SurgBleed} \\
    \cline{2-3}
        & \makecell[c]{mIoU $\uparrow$} & \makecell[c]{mIoU $\uparrow$} \\
    \hline
        \makecell[c]{\cmark} & \makecell[c]{41.89} & \makecell[c]{69.31} \\ 
        \makecell[c]{\xmark} & \makecell[c]{42.11} & \makecell[c]{69.70} \\
    \hline \toprule 
    \end{tabular}}
    \vspace{-5mm}
    \label{tab:ablation-freeze}
\end{table}

\textbf{Feasibility of Deployment.} Previous work~\cite{2024Spike} has demonstrated the feasibility of deploying SNN models on neuromorphic hardware, achieving even lower latency and energy consumption than theoretical estimates in Tab.~\ref{tab:endovis18} and~\ref{tab:surgbleed}. Given that our architecture shares fundamental similarities with the model presented in~\cite{2024Spike}, the deployment of our SNN framework on neuromorphic hardware is expected to be equally feasible. Nevertheless, as highlighted in a recent \textit{Nature} review~\cite{2025Neuromorphic} and analysis~\cite{neuromorphic_commercial_success}, the neuromorphic hardware ecosystem remains in an early developmental stage, particularly in supporting large-scale spike-driven model training and optimization. Such asymmetry is typical in emerging computing paradigms, where algorithmic advances often precede the maturation of corresponding hardware platforms. Overall, these trends indicate a promising yet still emerging pathway toward neuromorphic deployment of our model.


\beforesection
\section{Conclusion}
\aftersection
\revision{
We presented \textit{SpikeSurgSeg}, the first SNN-based framework for surgical scene segmentation. By combining a spike-driven video Transformer with tailored spike-informed pretraining, the framework achieves segmentation accuracy comparable to most ANN baselines while reducing inference latency and energy consumption by over an order of magnitude. In particular, spike-informed MAE pretraining improves spatiotemporal representation learning for sparse spike-driven features, while multi-spectral knowledge distillation enhances semantic supervision by aligning ANN and SNN representations in the frequency domain. Together with a lightweight spike-driven segmentation head, these designs enable temporally consistent surgical video segmentation while preserving the low-latency and energy-efficient advantages of spike-driven inference. Overall, \textit{SpikeSurgSeg} offers a practical balance between accuracy and efficiency for resource-constrained operative environments. In future work, we will explore larger-scale SNN pretraining and extend the framework to broader dense surgical video understanding tasks.}




\bibliographystyle{IEEEtran}
\bibliography{IEEEabrv}
\end{document}